\documentclass[conference]{IEEEtran}
\usepackage[latin9]{inputenc}
\usepackage{amsmath}
\usepackage{graphicx}
\usepackage{textcomp}
\usepackage{balance}
\usepackage{xcolor}
\usepackage{soul}

\newcommand{\FaceForensics}{\textit{DF}}
\newcommand{\FaceForensicsPP}{\textit{DFD}}
\newcommand{\Kaggle}{\textit{DFDC}}
\newcommand{\Celebdf}{\textit{CelebDF}}

\usepackage{booktabs}

\usepackage{cite}

\usepackage{amsmath}

\ifCLASSOPTIONcompsoc
  \usepackage[caption=false,font=normalsize,labelfont=sf,textfont=sf]{subfig}
\else
  \usepackage[caption=false,font=footnotesize]{subfig}
\fi

\usepackage{stfloats}

\usepackage{url}

\begin{document}
\title{Training Strategies and Data Augmentations\\in CNN-based DeepFake Video Detection}

\author{
\IEEEauthorblockN{Luca Bondi, Edoardo Daniele Cannas, Paolo Bestagini, Stefano Tubaro}
\IEEEauthorblockA{Dipartimento di Elettronica, Informazione e Bioingegneria \\ Politecnico di Milano, Piazza Leonardo da Vinci 32, 20133 Milano, Italy \\
}
}

\maketitle

\begin{figure}[b]
\vspace{-0.3cm}
\parbox{\hsize}{\em
WIFS`2020, December, 6-11, 2020, New York, USA.
XXX-X-XXXX-XXXX-X/XX/\$XX.XX \ \copyright 2020 IEEE.
}\end{figure}

\begin{abstract}
The fast and continuous growth in number and quality of deepfake videos calls for the development of reliable detection systems capable of automatically warning users on social media and on the Internet about the potential untruthfulness of such contents. While algorithms, software, and smartphone apps are getting better every day in generating manipulated videos and swapping faces, the accuracy of automated systems for face forgery detection in videos is still quite limited and generally biased toward the dataset used to design and train a specific detection system. In this paper we analyze how different training strategies and data augmentation techniques affect CNN-based deepfake detectors when training and testing on the same dataset or across different datasets.
\end{abstract}

\IEEEpeerreviewmaketitle

\section{Introduction}

As the number of techniques and algorithms to generate deepfake videos and swap faces grows rapidly, the effort of the forensic community is steering even more towards the development of reliable, robust, and automated deepfake detection methods.
Techniques and pipelines for facial manipulation~\cite{Zollhofer2018} and facial expression transfer between videos~\cite{Thies2016face2face, Thies2019deferred} are rapidly improving~\cite{Li_2020_CVPR}, while the availability of source code (Deepfake~\cite{Deepfake}, FaceSwap~\cite{Faceswap}) and even smartphone apps (Impressions~\cite{Impressions}, Doublicat~\cite{Doublicat}) makes face swapping available to a wider audience with either legitimate or harmful intents.
Tampered video detection is not a novel task to the forensics community~\cite{Rocha2011a, Milani2012a, Stamm2013}. Codec history~\cite{bestagini2016codec, Padin2020}, copy-move  detection~\cite{Bestagini2013, Damiano2019}, frame duplication or deletion~\cite{Stamm2012, Gironi2014} are just a few examples of the many contributions in the last decades. The main drawback of the earlier systems developed by the community is that the exploited traces are inherently subtle and vanish with compression or multiple editing operations~\cite{Milani2012a}.
The first generation of deepfake detection methods exploited several semantic traces, including eye blinking~\cite{Li2018}, face warping~\cite{li2019exposing}, head poses~\cite{Yang2019} or lighting inconsistencies~\cite{matern2019exploiting}. Due to the improvement of new and more accurate generation techniques, methods based on semantic artifacts began to fail, leading to the proposal of data-driven solutions capable of providing localization information through multi-task learning~\cite{Nguyen2019}, attention mechanisms~\cite{dang2019detection}, and ensembles of CNN~\cite{Bonettini2020}.

As detecting manipulated faces in videos becomes more important~\cite{Agarwal_2019_CVPR_Workshops,verdoliva2020media}, many deepfake detection systems proposed in the literature and in challenges are based on 
data-driven approaches, often backed by one or more CNNs trained on a specific dataset. However, the black-box model of data-driven CNN-based methods is notoriously prone to a drawback: over-fitting. Oftentimes, a bare train/validation/test  split done within a single dataset collected with a uniform methodology and by a single team proves insufficient in avoiding over-fitting on that very same dataset conditions and scenarios.
A recent example is shown in \cite{DFDC_results}, where the winning model of the Facebook/Kaggle DeepFake Detection Challenge~\cite{Dolhansky2020DFDCKaggle} scored an Average Precision of $82.56\%$ on the public dataset used for the temporary leader board of the challenge, and then dropped to $65.18\%$ on the sequestered dataset used for the final evaluation. Moreover, it is known that data dependency creates the risk of developing solutions unable to generalize over unseen methods or contexts. 

While most detectors prove to be very effective on a test subset coming from the same data distribution they are trained on, what are the detection performance in a cross-dataset scenario? What happens when a CNN trained for deepfake detection on a dataset A is tested on dataset B, C, and D? 
As it is difficult to gain direct insights about what happens inside a CNN black-box model, in this paper we offer a set of preliminary analysis on cross-dataset performance of CNN-based deepfake detection approaches.
Rather than focusing on developing a new technique optimized for a specific dataset, we train one of the most popular architectures used by competitors in the DeepFake Detection Challenge\cite{Dolhansky2020DFDCKaggle} and we evaluate how different training approaches~\cite{Bonettini2020} and data augmentation techniques~\cite{wang2019cnngenerated} affect the intra-dataset and cross-dataset detection performances. We base our experiments on publicly accessible datasets, i.e., FaceForensics++~\cite{roessler2019faceforensics}, the DeepFake Detection Challenge Dataset~\cite{Dolhansky2020DFDCKaggle}, and CelebDF(v2)~\cite{Celeb_DF_cvpr20}. We focus on faces extracted from deepfake videos rather than just deepfake images, as video compression is usually stronger than image compression. We also perform some analysis taking into account a limited availability of training data.
Far from being an exhaustive evaluation or overview of all the available techniques and datasets, we wish to share with the readers some insights to consider when developing a new deepfake detection system.

\section{Methodology}
\label{methodology}

In order to effectively compare the intra-dataset and cross-dataset detection performances, we first need to define a homogeneous training and testing methodology.
The process of determining whether a face in a video is manipulated starts with a face detection and extraction phase. We rely on BlazeFace~\cite{blazeface2019}, a fast and GPU-enabled face detector, and we extract the face with the highest confidence from $32$ frames for each video, uniformly sampled over time. This choice follows from~\cite{Bonettini2020}, thus taking into account that time and computational power may be a limited resource. As the extracted faces have different scales and aspect ratios, we crop the faces with a fixed aspect ratio of 1:1 before resizing to a fixed size of $256\times256$ pixels.
Once faces are extracted and uniform in size, we train an EfficientNetB4~\cite{Tan2019efficientnet} architecture as reference CNN, due to its popularity in the DeepFake Detection Challenge. The trained model is used to predict the likelihood of each face being fake. Results are reported at frame level as the Area Under Curve (AUC) of a Receiver-Operating-Characteristic (ROC) curve.

Among the several available datasets, we select the following four, due to their availability and ease of access and download:
\begin{itemize}
    \item \FaceForensics{}: FaceForensics~\cite{roessler2019faceforensics}, in its original version with 1000 real videos and 4000 fake videos generated with four different methods.
    \item \FaceForensicsPP{}: Actors-based videos added to FaceForensics~\cite{FaceForensicsPlusPlus}, with 363 real and 3068 fake videos.
    \item \Kaggle{}: The DeepFake Detection Challenge~\cite{Dolhansky2020DFDCKaggle}, with 19154 real and 100000 fake videos.
    \item \Celebdf{}: The Celeb-DF(v2) dataset~\cite{Celeb_DF_cvpr20}, with 890 real and 5639 fake videos.
\end{itemize}
The four dataset are divided into disjoint train, validation, and test sets at video level. In particular, for \FaceForensics{} and \FaceForensicsPP{} we follow the $720/140/140$ split proportion as suggested in~\cite{roessler2019faceforensics}. For \Kaggle{} we use the folders from $40$ to $49$ as test set and the folders from $35$ to $39$ as validation set. The remaining $40$ folders are the training set. For \Celebdf{} we use the test set provided by the dataset itself, and we randomly select $15\%$ of the videos as validation set, with the remaining $85\%$ for training. For both \FaceForensics{} and \FaceForensicsPP{} we consider only the videos compressed with H.264 at CRF $23$.

We run all our experiments with the PyTorch~\cite{pytorch} framework on a workstation equipped with two Intel Xeon E5-2687W-v4 and several NVIDIA Titan V.

\section{Baseline}
\label{baseline}

\begin{table}
\centering
\caption{ROC AUC for baseline intra and cross-dataset detection performance.}
\label{tab:fullDbResults}
\begin{tabular}{lrrrr}
\toprule
\textbf{Train\textbackslash Test} &  \textbf{CelebDF} &    \textbf{DF} &   \textbf{DFD} &  \textbf{DFDC}\\
\midrule
\textbf{CelebDF} &    0.998 & 0.615 & 0.708 & 0.665 \\
\textbf{DF     } &    0.734 & 0.960 & 0.844 & 0.695 \\
\textbf{DFD    } &    0.754 & 0.636 & 0.987 & 0.669 \\
\textbf{DFDC   } &    0.755 & 0.722 & 0.891 & 0.922 \\
\bottomrule
\end{tabular}
\end{table}

As a baseline for the upcoming experiments, we first need to evaluate the deepfakes detection performance of EfficientNetB4 trained as a classifier using the Binary Cross Entropy (BCE) loss.
The network is initialized with a model pre-trained on ImageNet, batch of $32$ faces, Adam optimizer, initial learning rate of $10^{-4}$ multiplied by a factor $0.1$ after $2000$ batch iterations with no reduction in validation loss. The training ends when the learning rate falls below $10^{-8}$. The final model is the one at the iteration that minimizes the validation loss.
Training and validation batches are always balanced, with randomly selected equal amounts of real and fake faces. No data augmentation is performed at this stage.
We train four CNN models on the training sets of the four datasets, then test each model against the test set of each dataset.

Results are reported in Table~\ref{tab:fullDbResults}, where the header column denotes the training dataset, while the header row reports the test dataset.
Reading the table by rows, we observe how on \Celebdf{} and \FaceForensicsPP{} the intra-dataset detection is very accurate, with an AUC above $0.98$. This, however, is not reflected on cross-dataset performance, as the model trained on \Celebdf{} and tested on \FaceForensicsPP{} presents an AUC of just $0.708$ ($29\%$ gap compared to intra-dataset AUC), while the model trained on \FaceForensicsPP{} and tested on \Celebdf{} reaches an AUC of $0.754$ ($23\%$ gap). The model trained on \FaceForensics{} has a slightly lower AUC when tested on the same dataset ($0.960$) with a $12\%$ gap when tested on \FaceForensicsPP{}. \Kaggle{} is the dataset presenting the lowest intra-dataset AUC ($0.922$) being at the same time the one that generalizes better, with $3\%$, $17\%$, and $20\%$ gap to \FaceForensicsPP{}, \Celebdf{}, and \FaceForensics{}, respectively. 
The baseline results are in line with what expected from data-driven methods: the largest dataset (i.e., \Kaggle{}) seems to provide more variety during the training phase, thus better generalization on unseen data.

\section{Training strategy}
\label{training}

\begin{table}
\centering
\caption{ROC AUC for triplet training intra and cross-dataset detection performance.}
\label{tab:fullDbTripletResults}
\begin{tabular}{lrrrr}
\toprule
\textbf{Train\textbackslash Test} &  \textbf{CelebDF} &    \textbf{DF} &   \textbf{DFD} &  \textbf{DFDC}\\
\midrule
\textbf{CelebDF} &    0.995 & 0.557 & 0.554 & 0.619 \\
\textbf{DF     } &    0.717 & 0.960 & 0.829 & 0.684 \\
\textbf{DFD    } &    0.759 & 0.709 & 0.882 & 0.666 \\
\textbf{DFDC   } &    0.773 & 0.714 & 0.886 & 0.907 \\
\bottomrule
\end{tabular}
\end{table}

\begin{figure*}
\centering
\includegraphics[width=\linewidth]{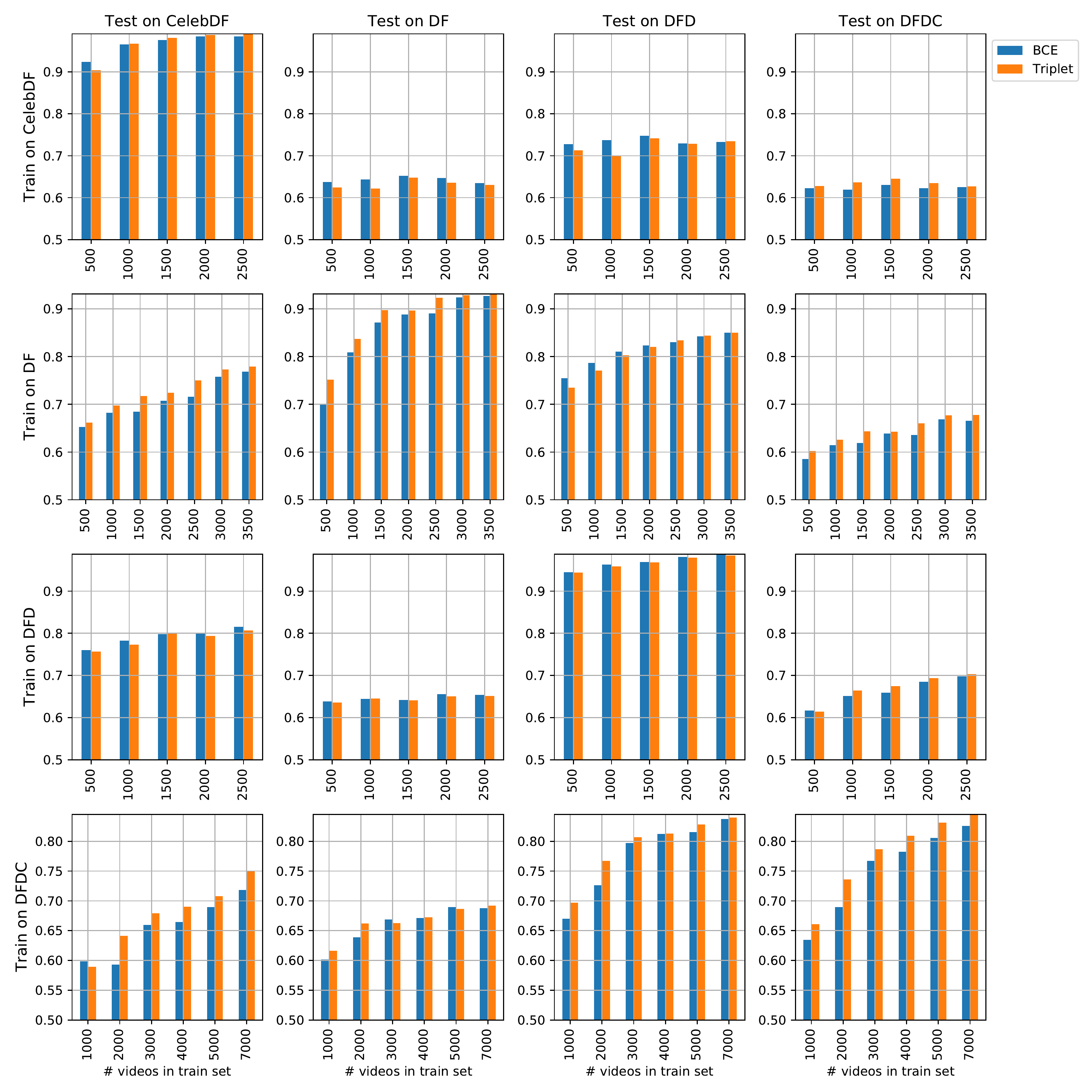}
\caption{ROC AUC for BCE and triplet training in data-limited conditions. For each dataset two CNNs are  trained selecting an increasing number of videos with BCE and triplet loss. Interestingly, we can see that the cross-dataset performances are generally higher on \FaceForensicsPP{}. This might be related to the overall quality of the dataset: while \FaceForensicsPP{} consists generally of high resolution videos, the other ones are more various and present also low quality samples. Training for detection in such difficult settings therefore might be helpful in generalizing on different, yet of higher quality, datasets.}
\label{fig:limited_data}
\end{figure*}

\begin{figure*}
\centering
\subfloat[Trained on \Celebdf{} with BCE loss]{\includegraphics[trim=0 0 0 23,clip,width=2.5in]{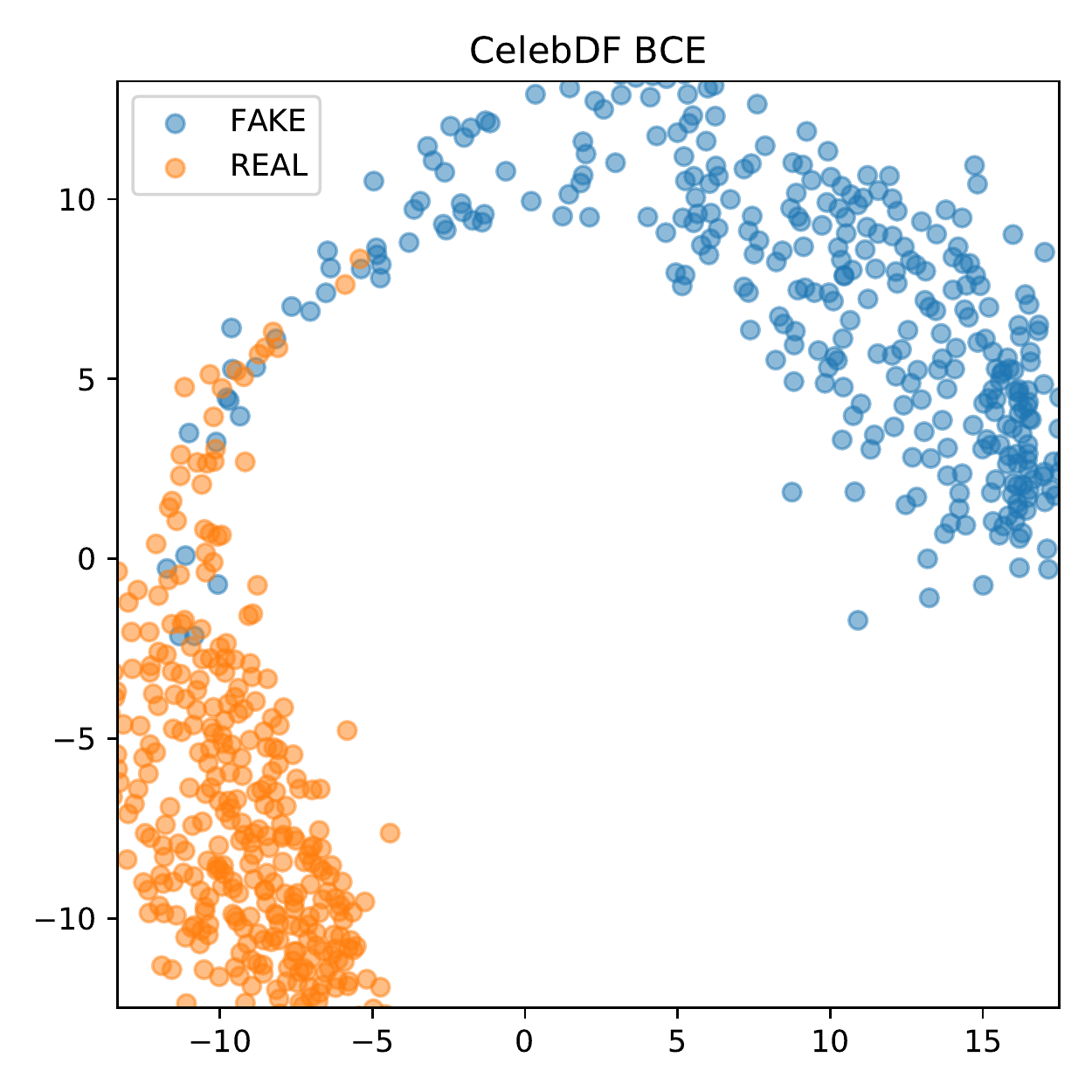}%
\label{fig:projection:celebdf_bce}}
\hfil
\subfloat[Trained on \Celebdf{} with triplet loss]{\includegraphics[trim=0 0 0 23,clip,width=2.5in]{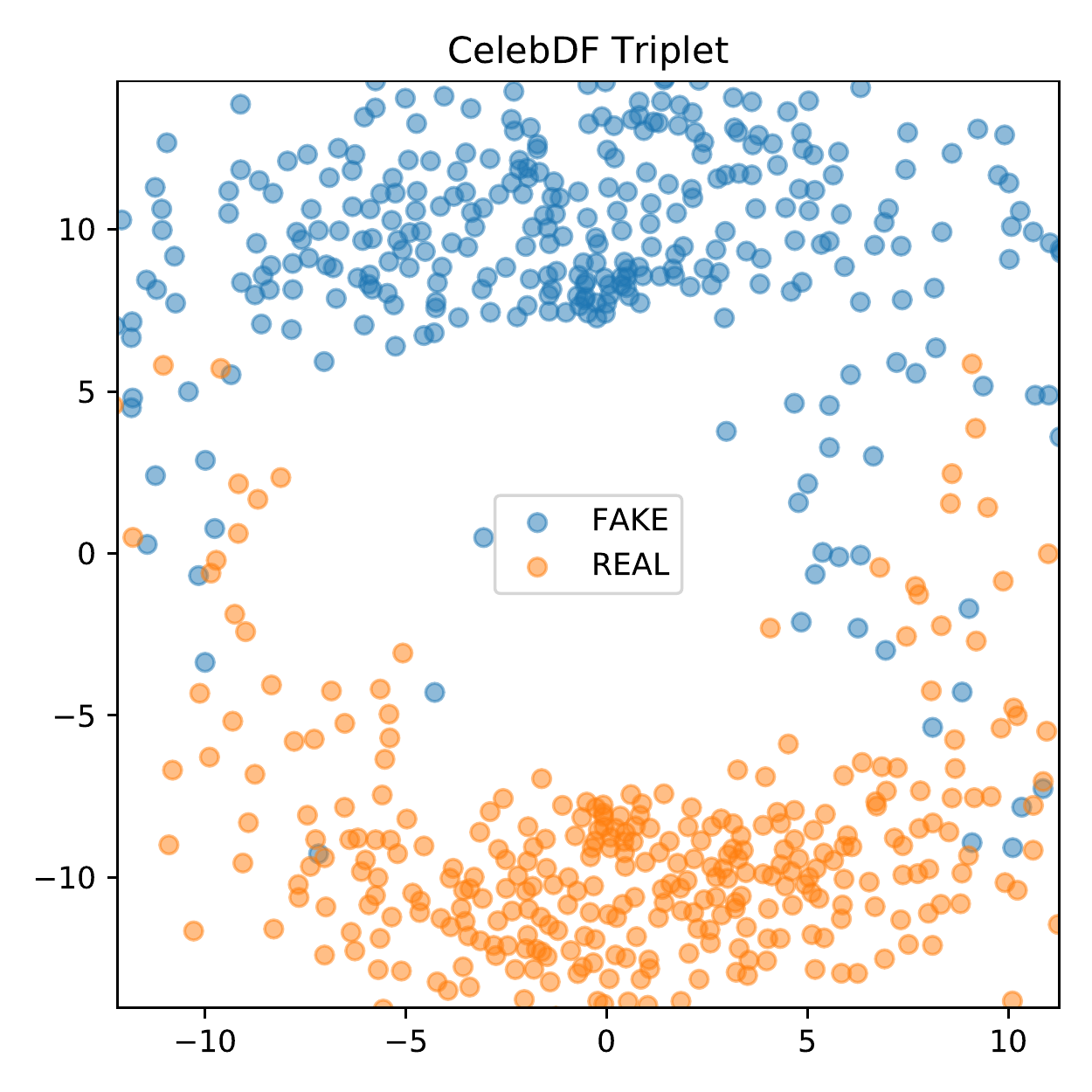}%
\label{fig:projection:celebdf_triplet}}

\subfloat[Trained on \Kaggle{} with BCE loss]{\includegraphics[trim=0 0 0 23,clip,width=2.5in]{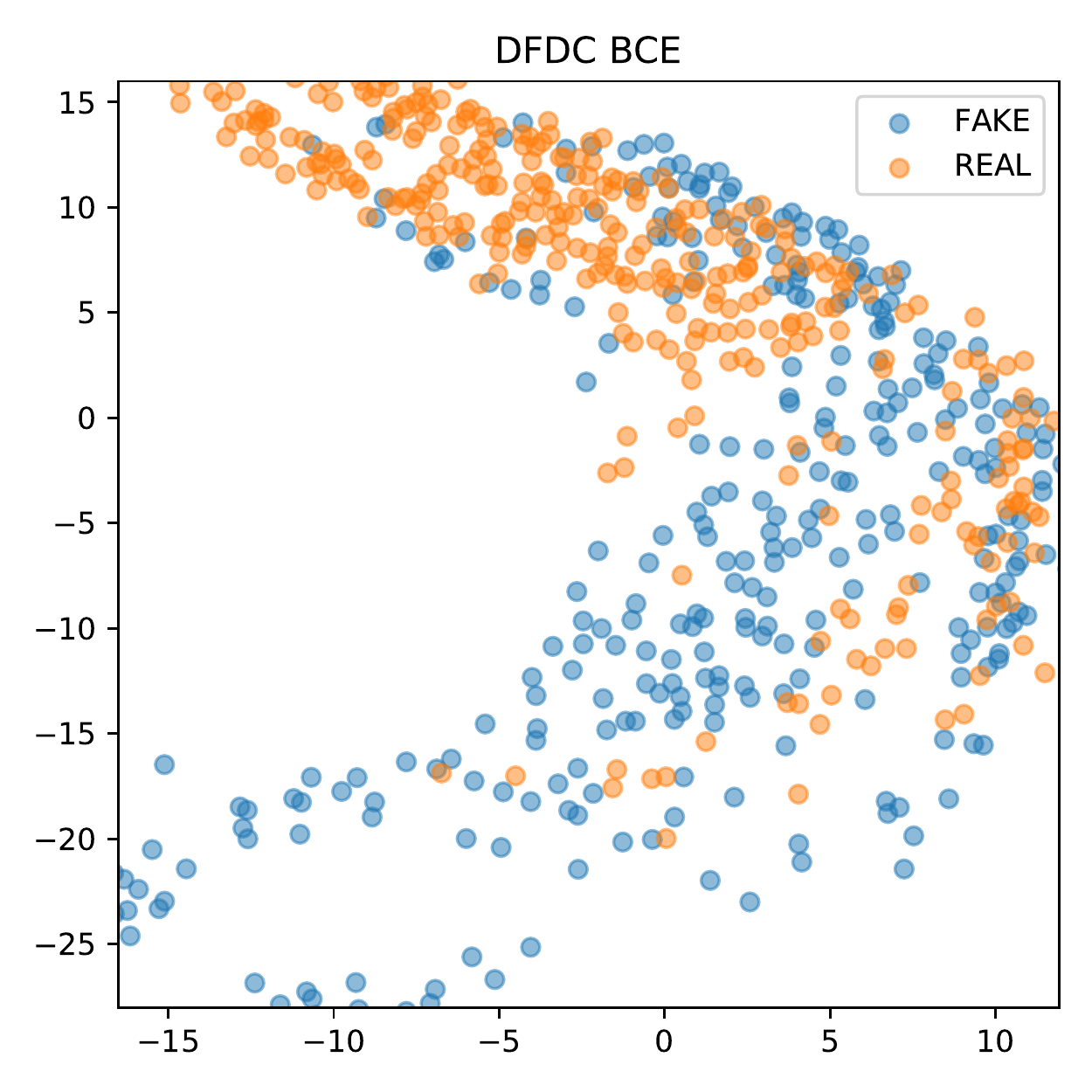}%
\label{fig:projection:dfdc_bce}}
\hfil
\subfloat[Trained on \Kaggle{} with triplet loss]{\includegraphics[trim=0 0 0 23,clip,width=2.5in]{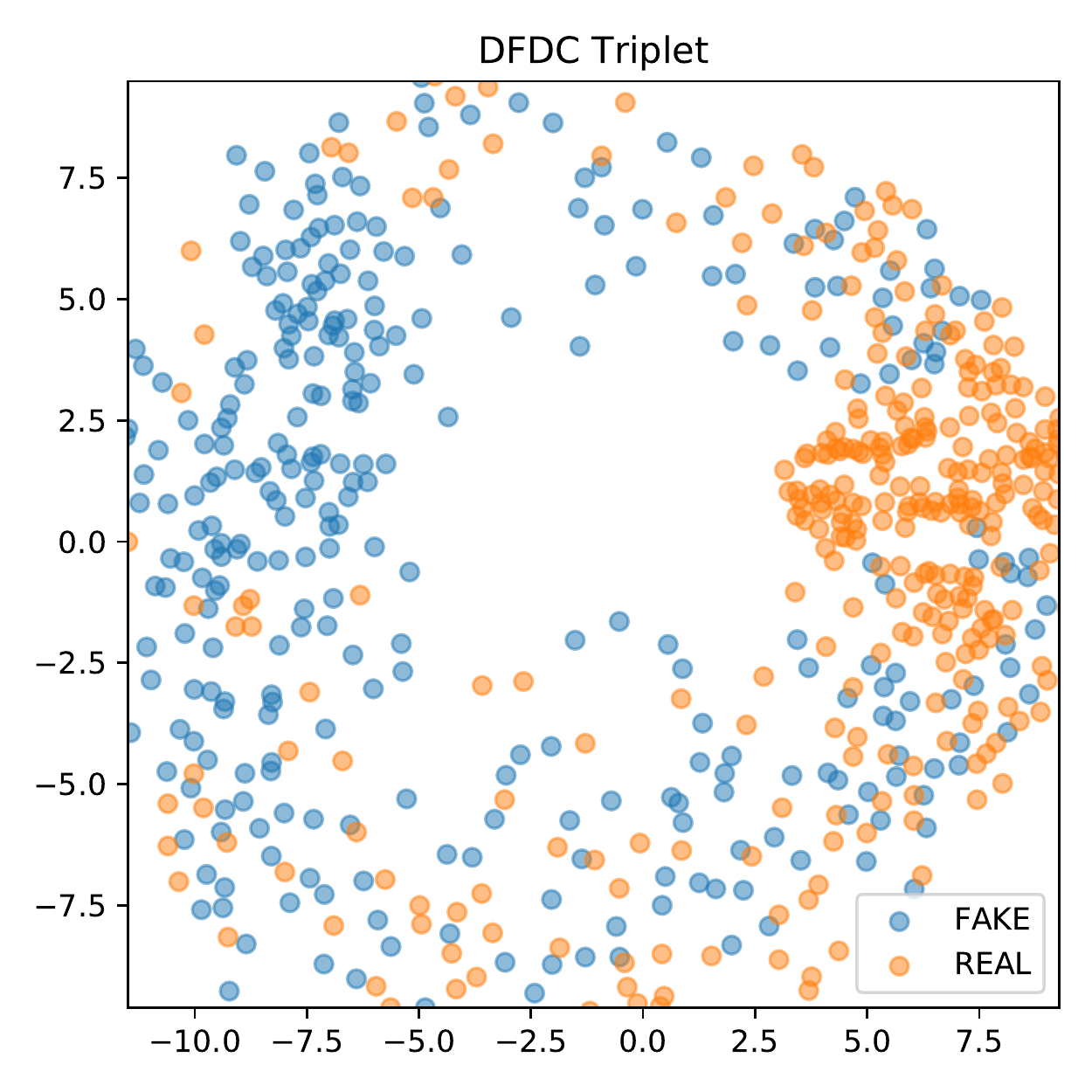}%
\label{fig:projection:dfdc_triplet}}

\caption{MDS projection of $10$ pairs of REAL/FAKE videos from the \Celebdf{} test dataset. Each point represent a frame in a video, $32$ frames are extracted from each video. Projections are produced starting from the features extracted by an EfficientNetB4 architecture trained on different datasets with binary cross entropy (BCE) or triplet loss.}
\label{fig:projection}
\end{figure*}

The first analysis we perform is related to the training strategy adopted for the CNN. Instead of relying on Binary Cross Entropy (BCE) loss, we train the CNN with a triplet loss~\cite{Wang2014triplet}, by running the CNN up to the last-minus-one layer (features layer). Considering triplets as (anchor sample, positive sample, negative sample), we generate the training triplets as (fake face, fake face, real face) and (real face, real face, fake face) in an equal number for each batch, so to balance the batch itself.
The training proceeds in a two step fashion.  

In the first step the CNN is trained with triplet loss up to the features layer. In the second step, only the last layer (classifier) of the CNN is trained (fine tuned) with binary cross entropy. 
In the context of the EfficientNetB4 architecture, feature vectors are $1792$ elements while the classifier has $1793$ weights ($1792$ multipliers and one bias coefficient). This means the classification layer accounts for less than $0.01\%$ of the net coefficients.
Triplet training is initialized with the model trained through BCE from the baseline, as this provides a faster convergence and prevents the model from failing into a trivial solution (all-zeros feature vector). The batch size is $10$ triplets to fit into $12$GB of GPU memory, the initial learning rate is set to $10^{-5}$ and it is dropped by a factor $10$ after $500$ batch iterations with no improvements on the validation loss. 

The fine tuning of the classifier is initialized with the triplet-trained model, with an initial learning rate of $10^{-6}$ dropped by a factor $10$ after $100$ iterations with no validation loss improvements. Both the triplet training and the fine-tuning process are stopped when the learning rate falls below $10^{-8}$. For both steps, the model at the iteration with the smallest validation loss is selected as the final one.

As for the baseline, we are interested in understanding both the intra and cross-dataset detection performance, as reported in Table~\ref{tab:fullDbTripletResults}. The results for \FaceForensics{}, \FaceForensicsPP{}, and \Celebdf{} show almost the same intra-detection AUC as with BCE training, with a loss in generalization capability more marked for the \Celebdf{} dataset. For the model trained on \Kaggle{}, the intra-detection AUC is similar to the BCE training, with slightly better cross-dataset AUC (with a modest $2\%$ increase in AUC with respect to the same combination in BCE training) only when testing on \Celebdf{}.

A different perspective on the differences between BCE and triplet losses is offered in Figure~\ref{fig:limited_data}, where EfficientNetB4 is trained in data-limited conditions by sub-sampling the training dataset. In this context, triplet loss proves beneficial in intra-dataset detection (\Kaggle{}, \Celebdf{}, and \FaceForensics{}) as well as in cross-dataset detection and outperforms BCE.

Even though the triplet training procedure is not revolutionary in terms of AUC, we are interested in analyzing the differences in representations learned at feature level with BCE and triplet loss on the same dataset and across different datasets. To this end, Figure~\ref{fig:projection} shows the Multidimensional Scaling (MDS) projection on two components of the features extracted with four differently trained EfficientNetB4 models. All four subplots project faces from the very same $10$ pairs of real/fake videos randomly extracted from the \Celebdf{} test set. Figure~\ref{fig:projection:celebdf_bce} uses features extracted with the CNN trained on the \Celebdf{} dataset with BCE, while Figure~\ref{fig:projection:celebdf_triplet} uses features extracted with the CNN trained on the same dataset with triplet loss instead. While in both cases the separation between real and fake frames is quite evident, in the triplet case the overlapping frames are less in number. This improvement could prove useful when aggregating the predictions from several frames at video level. 
Figure~\ref{fig:projection:dfdc_bce} and~\ref{fig:projection:dfdc_triplet} are generated with features extracted by CNNs trained on \Kaggle{} with BCE and triplet loss respectively. While certainly the overlap between real and fake frames is more evident than in Figures~\ref{fig:projection:celebdf_bce} and~\ref{fig:projection:celebdf_triplet}, the triplet loss seems to offer a bit more separation between the two classes, despite the feature extractor being trained on a different dataset.

\section{Data augmentation}
\label{augmentations}

\begin{figure*}
\centering
\includegraphics[width=\linewidth,trim=0 5 0 0,clip]{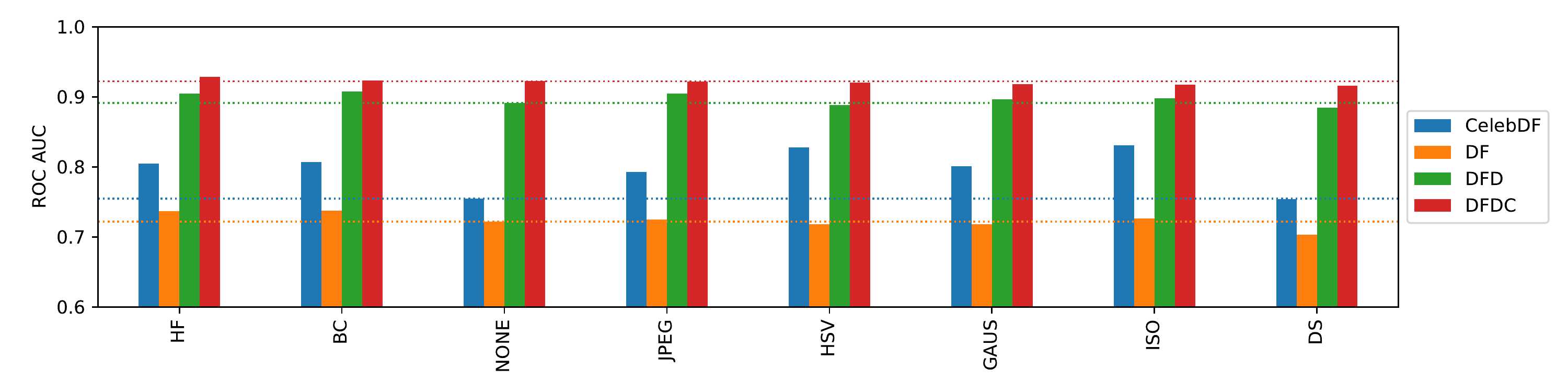}
\caption{ROC AUC of EfficientNetB4 trained with BCE on \Kaggle{} with different augmentation techniques. HF: Horizontal Flip. BC: Brightness and Contrast change. HSV: Hue, Saturation and Value changes. ISO: Addition of ISO noise. GAUS: Addition of Gaussian noise. DS: Down-scaling. JPEG: JPEG compression. MIX: baseline mix of all the other single augmentations. NONE: no augmentations. The horizontal lines are the AUC values when no augmentations are used.}
\label{fig:augmentations}
\vspace{-1em}
\end{figure*}

\begin{table}
\centering
\caption{ROC AUC for EfficientNetB4 trained with BCE and selected augmentations.}
\label{tab:fullDbAugResults}
\begin{tabular}{lrrrr}
\toprule
\textbf{Train\textbackslash Test} &  \textbf{CelebDF} &    \textbf{DF} &   \textbf{DFD} &  \textbf{DFDC} \\
\midrule
\textbf{CelebDF} &    0.998 & 0.616 & 0.795 & 0.673 \\
\textbf{DF     } &    0.764 & 0.966 & 0.847 & 0.691 \\
\textbf{DFD    } &    0.842 & 0.650 & 0.990 & 0.690 \\
\textbf{DFDC   } &    0.826 & 0.733 & 0.923 & 0.919 \\
\bottomrule
\end{tabular}
\end{table}

\begin{table}
\centering
\caption{ROC AUC for for EfficientNetB4 trained with triplet loss and selected augmentations.}
\label{tab:fullDbAugTripletResults}
\begin{tabular}{lrrrr}
\toprule
\textbf{Train\textbackslash Test} &  \textbf{CelebDF} &    \textbf{DF} &   \textbf{DFD} &  \textbf{DFDC} \\
\midrule
\textbf{CelebDF} &    0.995 & 0.570 & 0.604 & 0.595 \\
\textbf{DF     } &    0.779 & 0.963 & 0.858 & 0.682 \\
\textbf{DFD    } &    0.809 & 0.658 & 0.982 & 0.694 \\
\textbf{DFDC   } &    0.777 & 0.725 & 0.905 & 0.889 \\
\bottomrule
\end{tabular}
\end{table}

The second batch of experiments is devoted to understanding the effect of different data augmentation techniques. It is known that for deepfake images~\cite{wang2019cnngenerated} some type of data augmentation techniques prove beneficial in terms of robustness and cross-dataset generalization.
Among the many possible data augmentations techniques, we focus on the subset that could represent the transformations a face undergoes in the wild. The following augmentations are considered:
\begin{itemize}
    \item HF: Horizontal Flip
    \item BC: Brightness and Contrast changes
    \item HSV: Hue, Saturation and Value changes
    \item ISO: Addition of ISO noise
    \item GAUS: Addition of gaussian noise
    \item DS: Downscaling with a factor between $0.7$ and $0.9$
    \item JPEG: JPEG compression with a random quality factor between $50$ and $99$
\end{itemize}
We test the aforementioned augmentations independently, training with BCE on the \Kaggle{} dataset. All the proposed experiments are performed with the Albumentations~\cite{albumentations} framework. Results are reported in Figure~\ref{fig:augmentations}, ordered left to right in decreasing order of AUC on the \Kaggle{} test set. Two interesting considerations can be drawn in light of these results.
 
First, augmentations do not seem to help much increasing intra-dataset detection, maybe due to the cross-contamination between train, validation, and test set in terms of video settings and scenarios. The only exception is the HF augmentation, that provides a boost of just $0.7\%$ in AUC.
 
Second, some augmentations are beneficial (at times by a large margin) in terms of cross-dataset generalization. In particular, HF, BC, HSV, and JPEG provide for an AUC increase on networks trained on both \Celebdf{} and \FaceForensicsPP{}.
 
\FaceForensics{} does not seem to benefit much from augmentations, maybe due to the very different scenes depicted in \Kaggle{} compared to the ones in \FaceForensics{}. While the former has actors at distance, moving in the scene, often two actors, the latter has almost only a single actor, in the center of the scene, in a TV studio or during an interview with studio-level lights. 

In light of the results in terms of single augmentation, we build a data augmentation pipeline based on HF, BC, HSV, and JPEG, and re-train the CNN with both BCE and triplet loss. Table~\ref{tab:fullDbAugResults} reports results for BCE loss. The fusion of augmentations brings important improvements in terms of cross-dataset detection AUC, with up to $+9\%$ when training on \FaceForensicsPP{} and testing on \Celebdf{}, and when training on \Celebdf{} and testing on \FaceForensicsPP{}. The intra-dataset detection performances are instead mostly unaffected.
With augmentations applied to the CNN trained with triplet loss, Table~\ref{tab:fullDbAugTripletResults} shows how the few beneficial effects of triplet loss when training on full dataset are not visible anymore. In facts, triplet loss with data augmentations provides lower AUC for almost all combinations compared to BCE loss with data augmentation.

\section{Conclusions}
\label{conclusion}

Two are the main conclusions we can draw from the experiments presented in this paper.
First, a carefully built and tested data-augmentation pipeline can prove useful in increasing the generalization of a CNN model for deepfake video detection across different datasets. Not all augmentations are beneficial though, and checking the usefulness of each type of augmentation could be an important step in the workflow of developing a detection pipeline.
Second, triplet loss proves to be helpful in terms of both intra-dataset and cross-dataset detection performances under limited availability of training data. When large datasets are available, data augmentation on a BCE-trained CNN architecture proves to be the winning combination. 

\balance

\bibliographystyle{IEEEtran}
\bibliography{bibliography}

\end{document}